\documentclass[10pt,logo,copyright]{giga-report}
\usepackage[authoryear,sort&compress,round]{natbib}

\usepackage[utf8]{inputenc} %
\usepackage[T1]{fontenc}    %

\usepackage{parskip}        %
\usepackage{url}            %
\usepackage{booktabs}       %
\usepackage{amsfonts}       %
\usepackage{nicefrac}       %
\usepackage{microtype}      %
\usepackage{xcolor}         %
\usepackage[dvipsnames]{xcolor} %
\usepackage{graphicx}
\usepackage{animate}        %
\usepackage{subcaption}
\usepackage{tabularx}
\usepackage{makecell}
\usepackage{adjustbox}
\usepackage{setspace}
\usepackage{todonotes}
\usepackage{colortbl}
\usepackage{wrapfig}

\newcolumntype{M}[1]{>{\centering\arraybackslash}m{#1}}
\usepackage{float}
\usepackage{tikz}
\usetikzlibrary{positioning,shapes,arrows}
\usepackage{amsmath,amsfonts,bm, bbm,leftindex}
\usepackage{multirow}
\usepackage{comment}
\usepackage{gensymb}
\usepackage{lipsum}
\usetikzlibrary{arrows.meta, positioning, fit}
\usepackage[para]{threeparttable}
\usepackage{tikz}
\usetikzlibrary{tikzmark}

\def\eqref#1{equation~\ref{#1}}

\def\1{\bm{1}}

\DeclareMathAlphabet{\mathsfit}{\encodingdefault}{\sfdefault}{m}{sl}
\SetMathAlphabet{\mathsfit}{bold}{\encodingdefault}{\sfdefault}{bx}{n}

\makeatletter
\let\save@mathaccent\mathaccent
\newcommand*\if@single[3]{%
  \setbox0\hbox{${\mathaccent"0362{#1}}^H$}%
  \setbox2\hbox{${\mathaccent"0362{\kern0pt#1}}^H$}%
  \ifdim\ht0=\ht2 #3\else #2\fi
  }
\newcommand*\rel@kern[1]{\kern#1\dimexpr\macc@kerna}
\newcommand*\widebar[1]{\@ifnextchar^{{\wide@bar{#1}{0}}}{\wide@bar{#1}{1}}}
\newcommand*\wide@bar[2]{\if@single{#1}{\wide@bar@{#1}{#2}{1}}{\wide@bar@{#1}{#2}{2}}}
\newcommand*\wide@bar@[3]{%
  \begingroup
  \def\mathaccent##1##2{%
    \let\mathaccent\save@mathaccent
    \if#32 \let\macc@nucleus\first@char \fi
    \setbox\z@\hbox{$\macc@style{\macc@nucleus}_{}$}%
    \setbox\tw@\hbox{$\macc@style{\macc@nucleus}{}_{}$}%
    \dimen@\wd\tw@
    \advance\dimen@-\wd\z@
    \divide\dimen@ 3
    \@tempdima\wd\tw@
    \advance\@tempdima-\scriptspace
    \divide\@tempdima 10
    \advance\dimen@-\@tempdima
    \ifdim\dimen@>\z@ \dimen@0pt\fi
    \rel@kern{0.6}\kern-\dimen@
    \if#31
      \overline{\rel@kern{-0.6}\kern\dimen@\macc@nucleus\rel@kern{0.4}\kern\dimen@}%
      \advance\dimen@0.4\dimexpr\macc@kerna
      \let\final@kern#2%
      \ifdim\dimen@<\z@ \let\final@kern1\fi
      \if\final@kern1 \kern-\dimen@\fi
    \else
      \overline{\rel@kern{-0.6}\kern\dimen@#1}%
    \fi
  }%
  \macc@depth\@ne
  \let\math@bgroup\@empty \let\math@egroup\macc@set@skewchar
  \mathsurround\z@ \frozen@everymath{\mathgroup\macc@group\relax}%
  \macc@set@skewchar\relax
  \let\mathaccentV\macc@nested@a
  \if#31
    \macc@nested@a\relax111{#1}%
  \else
    \def\gobble@till@marker##1\endmarker{}%
    \futurelet\first@char\gobble@till@marker#1\endmarker
    \ifcat\noexpand\first@char A\else
      \def\first@char{}%
    \fi
    \macc@nested@a\relax111{\first@char}%
  \fi
  \endgroup
}
\makeatother

\definecolor{darkred}{rgb}{0.7, 0.0, 0.0}

\usepackage{amsmath} 
\usepackage{dsfont}
\usepackage{bbm}

\usepackage[table]{xcolor}

\usepackage{pifont}
\usepackage{graphicx}

\usepackage[nameinlink]{cleveref}
\crefname{equation}{Eq.}{Eqs.}
\crefname{figure}{Fig.}{Figs.}
\crefname{section}{Sec.}{Sec.}
\crefname{appendix}{App.}{App.}
\crefname{table}{Tab.}{Tabs.}
\crefname{algorithm}{Algo}{Algo}
\crefname{thm}{Thm}{Thm}
\Crefname{thm}{Thm}{Thm}
\crefname{prop}{Prop}{Prop}

\newcommand{\crefnames}[3]{%
  \@for\next:=#1\do{%
    \expandafter\crefname\expandafter{\next}{#2}{#3}%
  }%
}

\title{GigaBrain-0.5M*: a VLA That Learns From World Model-Based Reinforcement Learning}

\author{
\vspace{-0.1in}
\centerline{GigaAI} 
\centerline{{Project Page: \href{https://gigabrain05m.github.io}{https://gigabrain05m.github.io}}} 
\footnotesize
\textbf{GigaBrain Team (alphabetical order)}:
\normalfont
  Boyuan Wang,
  Bohan Li,
  Chaojun Ni,
  Guan Huang,
  Guosheng Zhao,
  Hao Li,
  Jie Li,
  Jindi Lv, \\
  Jingyu Liu,
  Lv Feng,
  Mingming Yu,
  Peng Li,
  Qiuping Deng,
  Tianze Liu,
  Xinyu Zhou,
  Xinze Chen,
  Xiaofeng Wang, \\
  Yang Wang,
  Yifan Li, 
  Yifei Nie,
  Yilong Li,
  Yukun Zhou,
  Yun Ye,
  Zhichao Liu,
  Zheng Zhu
\vspace{-1em}
}

\begin{document}
\maketitle

\begin{center}
    \centering
    \captionsetup{type=figure, justification=justified, singlelinecheck=false}
    \includegraphics[width=1\linewidth]{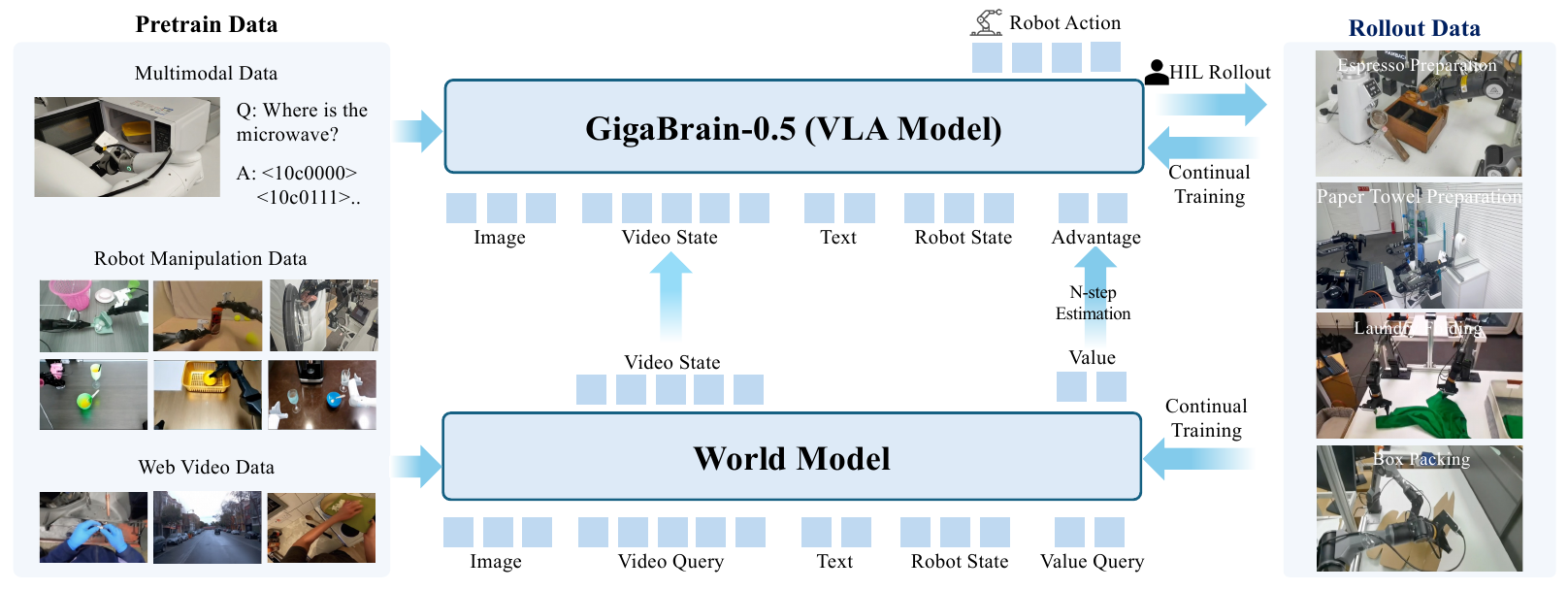}
\vspace{-0.2in}
    \caption{
\textit{GigaBrain-0.5M*} is a world model-conditioned VLA trained via world model-based reinforcement learning. Pretrained on multimodal, robot manipulation, and web video data, it enables self-improvement through human-in-the-loop (HIL) rollout that generates diverse training data for continual training.}
    \label{fig:teaser}
\end{center}

\begin{abstract}
\vspace{-0.1in}

Vision-language-action (VLA) models that directly predict multi-step action chunks from current observations face inherent limitations due to constrained scene understanding and weak future anticipation capabilities. In contrast, video world models pre-trained on web-scale video corpora exhibit robust spatiotemporal reasoning and accurate future prediction, making them a natural foundation for enhancing VLA learning. Therefore, we propose \textit{GigaBrain-0.5M*}, a VLA model trained via world model-based reinforcement learning. Built upon \textit{GigaBrain-0.5}, which is pre-trained on over 10,000 hours of robotic manipulation data, whose intermediate version currently ranks first on the international RoboChallenge benchmark. \textit{GigaBrain-0.5M*} further integrates world model-based reinforcement learning via \textit{RAMP} (Reinforcement leArning via world Model-conditioned Policy) to enable robust cross-task adaptation. Empirical results demonstrate that \textit{RAMP} achieves substantial performance gains over the RECAP baseline, yielding improvements of approximately 30\% on challenging tasks including \texttt{Laundry Folding}, \texttt{Box Packing}, and \texttt{Espresso Preparation}. Critically, \textit{GigaBrain-0.5M$^*$} exhibits reliable long-horizon execution, consistently accomplishing complex manipulation tasks without failure as validated by real-world deployment videos on our \href{https://gigabrain05m.github.io}{project page}.

\end{abstract}

\abscontent
\section{Introduction}

Recent advances in vision-language-action (VLA) models~\citep{pi0,pi05,go1,gr00t,gr3,walloss,galaxea,gigabrain0} have demonstrated compelling results in understanding instructions, perceiving environments, and executing complex manipulation. Nevertheless, a fundamental limitation persists in mainstream VLA architectures: their reliance on myopic observations for long-horizon action planning, this shortcoming stems from an architectural bias toward reactive control rather than prospective planning. Conversely, foundation world models trained on massive-scale video corpora have demonstrated remarkable proficiency in forecasting plausible future states, such predictive priors offer a pathway to endow VLAs with foresight.

Therefore, we introduce \textit{GigaBrain-0.5M*}, a VLA model trained via world model-based reinforcement learning. 
Specifically, \textit{GigaBrain-0.5M*} extends \textit{GigaBrain-0.5M} (our latest VLA pre-trained on over 10K hours of real-world robotic interaction data) with \textit{RAMP} (Reinforcement leArning via world Model-conditioned Policy). The \textit{RAMP} pipeline operates in four iterative stages: (1) The world model is pretrained with large-scale robot manipulation data to predict value and future states. (2) The policy is fine-tuned by conditioning actions on the world model's predicted value and future state. (3) Deploy the policy in real environments, producing robot rollout data with human-in-the-loop intervention. (4) Continual training world model and policy with rollout data. This iterative training paradigm enables self-improvement.

Therefore, we introduce \textit{GigaBrain-0.5M*}, a VLA model trained via world model-based reinforcement learning. Specifically, \textit{GigaBrain-0.5M*} extends \textit{GigaBrain-0.5M} (our latest VLA pretrained on over 10K hours of real-world robotic interaction data) by integrating \textit{RAMP} (Reinforcement leArning via world Model-conditioned Policy). The \textit{RAMP} framework follows an iterative four-stage training paradigm. First, the world model is pretrained on large-scale robot manipulation data to forecast future states and associated value. Second, the policy undergoes fine-tuning by conditioning its action selection on the world model's predicted futures and value estimates. Third, the conditioned policy is deployed in physical environments to collect rollout trajectories under human-in-the-loop intervention. Fourth, both the world model and policy are jointly refined using the curated rollout dataset. This iterative training paradigm enables continual learning and self-improvement. The proposed \textit{RAMP} is inspired by RECAP in $\pi^{*}_{0.6}$~\citep{pi06}, as both approaches utilize additional information as conditions for the VLA model. However, RECAP only uses sparse advantages (0 or 1) as input, providing limited information gain. In contrast, our proposed \textit{RAMP} leverages future states predicted by a well-pretrained world model, yielding substantial information gain. Furthermore, we theoretically verify that RECAP is a special case of \textit{RAMP}.

In our experiments, we first conduct comprehensive internal evaluations to assess the performance of \textit{GigaBrain-0.5} against strong baselines, including $\pi_{0.5}$~\citep{pi05} and {GigaBrain-0}~\citep{gigabrain0}. Our method achieves state-of-the-art success rates across a diverse suite of manipulation tasks, with particularly pronounced advantages on challenging deformable object manipulation and long-horizon procedural tasks. Furthermore, an intermediate version of \textit{GigaBrain-0.5} secured the top position on the public RoboChallenge benchmark leaderboard~\citep{robochallenge}.
We additionally perform extensive ablation studies to analyze the impact of different reinforcement learning algorithms on real-robot performance. Results demonstrate that our proposed \textit{RAMP} significantly outperforms alternative approaches such as AWR~\citep{awr} and RECAP~\citep{pi06}, yielding superior multi-task generalization and markedly improved sample efficiency during policy learning.
Notably, \textit{GigaBrain-0.5M*} exhibits robust long-horizon reasoning capabilities, seamlessly executing complex sequential tasks, including laundry folding, box packing and espresso preparation, without interruption over extended interaction horizons.
\section{Related Works}

\subsection{Vision-Language-Action Models}

Recent progress in foundation language models has catalyzed the development of VLA models~\citep{o2024open, team2024octo, kimopenvla, pi0, pertsch2025fast, pi05, doshi2024scaling, wang2024scaling, liu2024rdt, qu2025spatialvla, li2024cogact, gr00t, gr3, gigabrain0, swiftvla}, which pursue enhanced cross-task and cross-embodiment generalization by jointly scaling model parameters and training corpora. Such systems commonly leverage frozen or fine-tuned vision-language backbones~\citep{qwen25vl, paligemma, kosmos, llava, flamingo, paligemma2, smolvlm} to process heterogeneous sensory inputs and produce executable motor commands, employing either autoregressive tokenization strategies or continuous action spaces formulated through flow-based generative paradigms~\citep{lipman2022flow, liu2022rectified}.
Although contemporary VLAs incorporate extensive cross-embodiment datasets~\citep{o2024open, khazatsky2024droid, dasari2019robonet, walke2023bridgedata, ebert2021bridge} alongside large-scale proprietary archives to bolster generalization capabilities, a fundamental limitation persists in their capacity for temporally extended reasoning. Specifically, these models exhibit a tendency to condition action generation predominantly on immediate observation inputs when addressing long-horizon manipulation tasks.

\subsection{World Models for Policy Models}
Recent breakthroughs in world modeling~\citep{hunyuanvideo,wan,cosmos,cosmospredict,vjepa2,genieEnvisioner,enerverseac,dreamgen} have accelerated the adoption of generated data to bridge the simulation-to-reality gap in embodied AI systems~\citep{sorasurvey}. In autonomous driving, world models are leveraged to generate corner cases data~\citep{drivedreamer,drivedreamer2,gaia,gaia2,magicdrive,vista,cosmosdrivedream} and construct traffic situations~\citep{,drivedreamer4d,recondreamer,recondreamer++,recondreamerrl}.
In embodied robotics contexts, techniques such as~\citep{gigaworld0} harness world-model-generated samples, spanning texture-varied scenes~\citep{emma,robotransfer,roboengine}, multi-viewpoint renderings~\citep{egodemogen}, and ego-centric translations~\citep{mimicdreamer}, to enrich the training data of VLA models. 
A distinct paradigm involves forecasting future visual trajectories via world models (e.g., DreamGen~\citep{dreamgen} and ViDAR~\cite{vidar}), subsequently inferring executable motor commands through Inverse Dynamics Models (IDMs). The efficacy of such pipelines critically hinges on the visual fidelity and physical plausibility of generated sequences.
Beyond data generation, emerging approaches investigate tighter integration between world models and policy learning. Methods like~\citep{worldvla,drivevla,drivedreamer,motus,lingbotva,mimicvideo} fuse latent representations from predictive world models with policy networks to improve sample efficiency and generalization. More ambitiously, frameworks such as~\citep{cosmospolicy} bypass explicit policy networks altogether, directly mapping world model predictions to action sequences.

\subsection{Reinforcement Learning for Vision-Language-Action Models}

Imitation learning policies suffer from compounding errors due to distribution shift~\cite{ross2011dagger}, inherently limiting their performance to the quality of the demonstration data. While DAgger and its variants~\cite{kelly2019hg,jang2022bc} mitigate this issue through online expert interventions, they still rely on continuous human supervision and lack mechanisms for autonomous policy improvement.
To transcend the limitations of imitation learning, reinforcement learning has been widely adopted for robotic policy optimization. Traditional approaches employ on-policy algorithms~\cite{schulman2017proximal} or off-policy methods~\cite{kalashnikov2018qt} to refine policies through environment interaction. Recent works extend these paradigms to VLA models via direct policy gradient optimization~\cite{Tan2025InteractivePostTraining,Lu2025VLARL} or residual policy learning on frozen backbones~\cite{Guo2025ImprovingVLA}. However, scaling policy gradient methods to large-scale VLAs remains challenging due to training instability and sample inefficiency.
An emerging direction circumvents explicit policy gradient computation by conditioning action generation on value signals, encompassing reward-conditioned policies~\cite{Kumar2019RewardConditionedPolicies}, and advantage-conditioned formulations~\cite{kuba2023advantage,Wu2023ElasticDecisionTransformer}. Recently, $\pi^*_{0.6}$~\cite{pi06} introduced the RECAP framework, demonstrating that advantage-conditioned reinforcement learning enables VLAs to attain high performance on downstream tasks through on-robot data collection. This motivates us to explore world model-based reinforcement learning, where a world model jointly predicts value and future states to serve as rich policy conditions.
\section{GigaBrain-0.5M*}

Building upon our foundation VLA model \textit{GigaBrain-0.5}, we introduce \textit{GigaBrain-0.5M*}, an enhanced policy model that integrates world model-based RL: \textit{RAMP} (Reinforcement leArning via world Model-conditioned Policy). This section first details the architecture and pre-training data composition of \textit{GigaBrain-0.5}. We then present \textit{RAMP}, a training methodology that harnesses world model predictions to iteratively refine policy behavior through experience and corrective feedback signals.

\subsection{GigaBrain-0.5}
\textit{GigaBrain-0.5} inherits the end-to-end VLA architecture of GigaBrain-0~\citep{gigabrain0}, designed to map visual observations and language instructions to action sequences for bi-manual robots. 
It employs a mixture-of-transformers~\citep{mot} backbone, utilizing a pre-trained PaliGemma-2~\citep{paligemma2} vision-language model (VLM) to encode multimodal inputs and an action Diffusion Transformer (DiT)~\citep{dit} with flow matching~\citep{lipman2022flow} to predict action chunks. 
To enhance reasoning capabilities, \textit{GigaBrain-0.5} generates an Embodied Chain-of-Thought (Embodied CoT) consisting of autoregressive subgoal language, discrete action tokens~\citep{pertsch2025fast}, and 2D manipulation trajectories $\mathbf{t}_{1:10}$. While language and discrete tokens are decoded via the VLM head, the 2D trajectory is regressed from learnable tokens through a lightweight GRU decoder. In this version, depth information and 2D trajectories are treated as optional states, allowing the model to adapt to varied sensor modalities and task requirements.
All components are jointly optimized under a unified objective:
\begin{equation}
    \mathcal{L} = \mathbb{E}_{\mathcal{D}, \tau, \epsilon}\left[
        -\sum_{j=1}^{n-1} M_{\text{CoT},j} \log p_{\theta}\left(x_{j+1} \mid x_{1:j}\right)
        + \left\| \epsilon - a_{\text{chunk}} - f_{\theta}\left(a_{\text{chunk}}^{\tau, \epsilon}\right) \right\|^{2}
        + \lambda \left\| \text{GRU}(\hat{\mathbf{t}}_{1:10}) - \mathbf{t}_{1:10} \right\|^2
    \right],
\end{equation}
where $\mathcal{D}$ is the training dataset and $M_{\text{CoT},j} \in \{0,1\}$ is a per-token mask indicating whether position $j$ belongs to the CoT reasoning stream (subgoal language or discrete actions). For the diffusion process, $\tau \in [0, 1]$ is the flow-matching timestep, $\epsilon \sim \mathcal{N}(0, \mathbf{I})$ is Gaussian noise, and $a_{\text{chunk}}^{\tau, \epsilon} = \tau \cdot a_{\text{chunk}} + (1 - \tau) \cdot \epsilon$ denotes the noised action chunk. The terms $\hat{\mathbf{t}}_{1:10}$ and $\mathbf{t}_{1:10}$ represent predicted and ground-truth trajectory keypoints, balanced by hyperparameter $\lambda$.
Notably, Knowledge Insulation~\citep{KI} inherently prevents optimization interference between the language and action prediction terms.

\begin{figure}[t]
\centering
\captionsetup{type=figure, justification=justified, singlelinecheck=false}
    \includegraphics[width=1\linewidth]{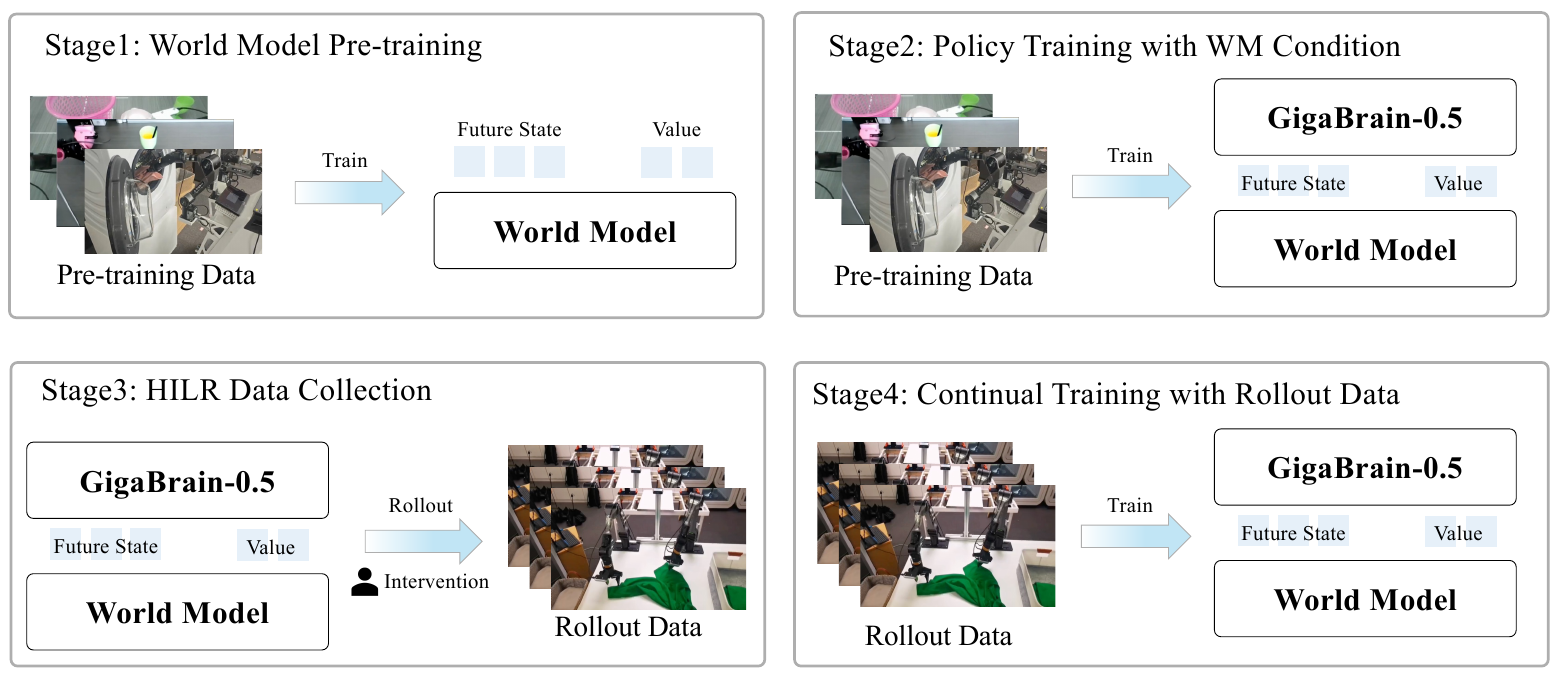}
    \caption{
    \textbf{Overview of \textit{RAMP}.} The \textit{RAMP} framework operates through a four-stage pipeline. (1) \textbf{World Model Pre-training} establishes a unified representation space for both future state prediction and value estimation. (2) \textbf{Policy Training with World Model Condition} initializes the \textit{GigaBrain-0.5} policy with explicit world model conditioning. (3) \textbf{Human-in-the-Loop Rollout (HILR) Data Collection} generates diverse and high-quality trajectories through autonomous execution followed by expert corrections. (4) \textbf{Continual Training with Rollout Data} updates the policy using the annotated trajectory data, incorporating both successful demonstrations and corrective signals. This tightly integrated closed-loop process facilitates continuous policy refinement and self-improvement.
    }
    \label{fig:RAMP}
\end{figure}

\subsection{RAMP}

In this section, we first formulate the proposed \textit{RAMP} framework and demonstrate that RECAP~\citep{pi06} is a special case within this formulation. Subsequently, we detail the implementation of \textit{RAMP}, which encompasses four iterative training stages: (1) World Model Pre-training, (2) Policy Pre-training,  (3) Human-in-the-Loop Rollout (HILR) Data Collection, and (4) Policy Training with Rollout Data.

\subsubsection{RAMP Formulation}
To derive a scalable training objective that leverages world model latents, we extend the KL-regularized reinforcement learning framework to our augmented state space $\mathbf{S} = (\mathbf{o}, \mathbf{z}, l)$,  where $\mathbf{z}$ represents the latent representation extracted by the world model. Our goal is to maximize expected returns while constraining the policy $\pi$ from deviating from a reference policy $\pi_{\text{ref}}(\cdot|\mathbf{S})$ via KL divergence. Drawing on standard results in regularized reinforcement learning, the closed-form solution for the optimal policy is given by~\citep{pi06}:
\begin{equation}
    \hat{\pi}(a|\mathbf{S}) \propto \pi_{\text{ref}}(a|\mathbf{S}) \exp\left( \frac{A^{\pi_{\text{ref}}}(\mathbf{S}, a)}{\beta} \right).
\end{equation}

To mitigate the numerical instability associated with directly estimating the exponential advantage term, we introduce a binary improvement indicator $I$ and assume that the probability of observing an improvement event $p(I|a, \mathbf{S})$ is proportional to the exponential advantage of the action. By applying Bayes' theorem, we reformulate this intractable advantage term as a ratio of conditional probabilities: $\exp(A^{\pi_{\text{ref}}}(\mathbf{S}, a) / \beta) \propto \pi_{\text{ref}}(a|I, \mathbf{S}) / \pi_{\text{ref}}(a|\mathbf{S})$. Substituting this ratio back into the optimal policy equation re-expresses $\hat{\pi}$ as a composition of the unconditional distribution and the conditional improvement distribution. Consequently, we parameterize a neural network $\pi_\theta$ to simultaneously fit these distributions, resulting in the final training objective of minimizing the weighted negative log-likelihood:

\begin{equation}
\label{eq:sft}
\mathcal{L}(\theta) = \mathbb{E}_{D} \left[ -\log \pi_\theta(a | \mathbf{o}, \mathbf{z}, l) - \alpha \log \pi_\theta(a | I, \mathbf{o}, \mathbf{z}_t, l) \right],
\end{equation}

where $I = \mathds{1}[A(\mathbf{o}, \mathbf{z}, l, a) > \epsilon]$ serves as the improvement signal. 

The explicit inclusion of the latent state $z$ in this objective is not merely a structural choice, but a theoretical necessity. To justify this design, we examine the relationship between our approach \textit{RAMP} and existing methods like RECAP~\citep{pi06} from a probabilistic perspective. We establish the intrinsic connection between these two paradigms.
From a probabilistic modeling perspective, we theoretically establish the intrinsic connection between \textit{RAMP} and RECAP, demonstrating that RECAP is essentially a degenerate special case of \textit{RAMP} where information about future latent states is ignored. Specifically, the policy form of RECAP, $\pi(a|o, I)$, is mathematically equivalent to the marginal distribution of the \textit{RAMP} policy $\pi(a|\mathbf{o}, \mathbf{z}, I)$ over the latent future state $\mathbf{z}$:

\begin{equation}
\pi_{RECAP}(a|\mathbf{o}, I) = \int_{z} \pi_{RAMP}(a|\mathbf{o}, \mathbf{z}, I) p(\mathbf{z}|\mathbf{o}, I) d\mathbf{z}.
\end{equation}

This implies that RECAP effectively learns an \textit{average policy} that must implicitly integrate over and compromise across all possible future evolutions without specific guidance. In contrast, \textit{RAMP} eliminates this uncertainty by explicitly conditioning on the world model's prediction $\mathbf{z}$, transforming the problem from an \textit{average guess} of the future into precise planning targeting a specific physical state.
Furthermore, from an information-theoretic standpoint, the introduction of the spatio-temporal latent $\mathbf{z}$ provides significant Information Gain for action generation. While RECAP relies solely on a sparse binary advantage signal ($I \in \{0, 1\}$) for coarse credit assignment, \textit{RAMP} leverages $\mathbf{z}$ to inject dense geometric structures and physical dynamics priors, thereby significantly reducing the conditional entropy of action generation $H(a|\mathbf{o}, \mathbf{z}, I) \le H(a|\mathbf{o}, I)$.

\subsubsection{The RAMP Implementation}

\textit{RAMP} enables VLA models to learn from experiences by incorporating world model guidance, throughout the entire training lifecycle. Spanning from large-scale offline pre-training to multi-round iterative fine-tuning on autonomous rollout data, our approach achieves progressive policy improvement. As illustrated in Fig.~\ref{fig:RAMP}, the pipeline is structured into four progressive stages:

\textbf{Stage 1: World Model Pre-training.} The initial phase establishes a world model $\mathcal{W}_\phi$ capable of jointly predicting future visual states and value estimates. Following the methodology of~\citet{pi06}, we derive sparse rewards from episode-level success labels such that the value function corresponds to the negative expected steps-to-completion. Specifically, the reward function is defined as:
\begin{equation}
    r_{t} = 
    \begin{cases}
        0 & \text{if } t = T \text{ and episode succeeds}, \\
        -C_{\text{fail}} & \text{if } t = T \text{ and episode fails}, \\
        -1 & \text{otherwise},
    \end{cases}
\end{equation}
where $T$ denotes the terminal timestep of the episode and $C_{\text{fail}}$ is a large positive constant chosen to ensure that failed episodes receive substantially lower cumulative returns than successful ones. This sparse reward formulation encourages policies to minimize execution time while prioritizing task completion over partial progress. Following the latent frame injection strategy~\citep{cosmospolicy}, we embed the value signal as an additional latent frame that is concatenated with the visual latent state before being fed into the world model. This approach requires no architectural modifications to the underlying Diffusion Transformer. Specifically, future visual observations $\{\mathbf{o}_{t+i}\}_{i \in \{12,24,36,48\}}$ are first encoded into spatiotemporal visual latents $\mathbf{z}_t \in \mathbb{R}^{H' \times W' \times C'}$ using a pre-trained VAE. Concurrently, scalar and low-dimensional auxiliary signals, including the current value estimate $v_t \in \mathbb{R}$ and proprioceptive state $\mathbf{p}_t \in \mathbb{R}^d$, are transformed via a spatial tiling projection $\Psi(\cdot)$. This projection replicates and broadcasts the low-dimensional vectors across the spatial dimensions to match the shape of the visual latents. The complete latent state is then constructed as:
\begin{equation}
    \mathbf{s}_t = \big[ \mathbf{z}_t \,;\, \Psi(v_t) \,;\, \Psi(\mathbf{p}_t) \big],
\end{equation}
where $[\cdot\,;\,\cdot]$ denotes channel-wise concatenation. This unified representation enables the world model to jointly reason about visual dynamics, task progress (via value), and robot kinematics within a single forward pass.

\begin{figure}[!t]
\centering
\captionsetup{type=figure, justification=justified, singlelinecheck=false}
\includegraphics[width=0.8\linewidth]{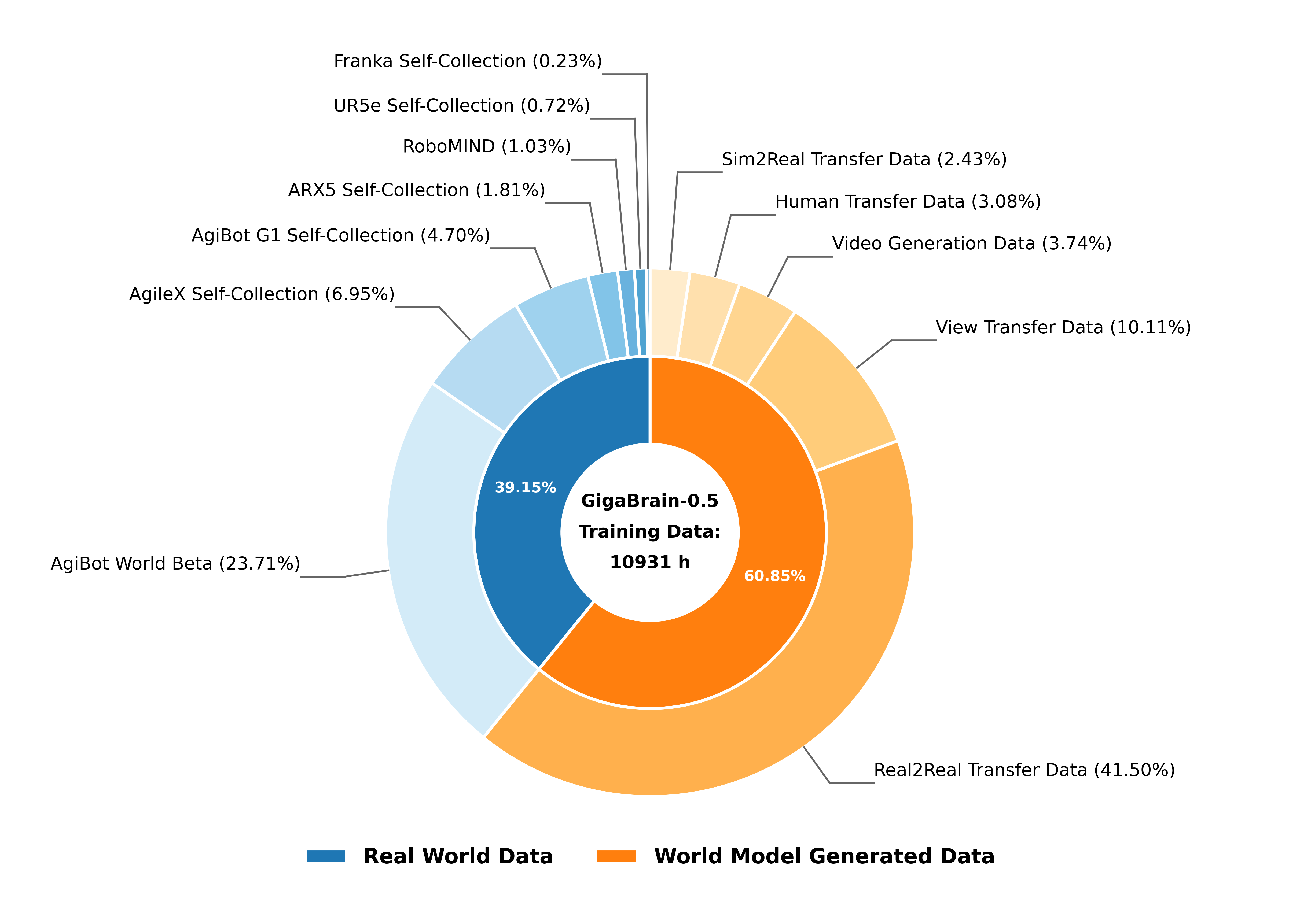}
\caption{Data distribution of the pre-training stage of \textit{GigaBrain-0.5}.}
\label{fig:pretraindata}
\end{figure}

We adopt Wan2.2~\citep{wan} as the backbone architecture for our world model $\mathcal{W}_\phi$. The model is trained via flow matching~\citep{lipman2022flow}. By treating future visual states and value estimates as temporally extended video frames, the DiT backbone naturally leverages its spatiotemporal self-attention mechanisms to model the relationship between current observations, actions, and future task outcomes:
\begin{equation}
\mathcal{L}_{\text{WM}} = \mathbb{E}_{\mathcal{D}, \tau, \epsilon}\left[
\left\| \mathcal{W}_\phi\big( \mathbf{s}_{\text{future}}^{\tau, \epsilon} \big) - (\mathbf{s}_{\text{future}} - \epsilon) \right\|^{2}
\right],
\end{equation}

\noindent where $\mathbf{s}_{\text{future}}^{\tau, \epsilon} = \tau \mathbf{s}_{\text{future}} + (1 - \tau) \epsilon$ denotes the linear interpolation between source noise $\epsilon \sim \mathcal{N}(0, \mathbf{I})$ and the ground-truth latent state sequence $\mathbf{s}_{\text{future}}$, with $\tau \sim \mathcal{U}(0,1)$. The target term $(\mathbf{s}_{\text{future}} - \epsilon)$ corresponds to the constant-velocity vector field along the Optimal Transport path between noise and data distributions. We utilize 4K hours real robot manipulation data to train the world model, the data distribution is visualized in Fig.~\ref{fig:pretraindata}.

\textbf{Stage 2: Policy Training with World Model Conditioning.} The second phase initializes the policy from the pretrained \textit{GigaBrain-0.5} checkpoint and further fine-tunes it with world model conditioning. Training details are provided in Sec.~\ref{sec:exp_RAMP}. Specifically, the policy receives two auxiliary signals predicted by the world model $\mathcal{W}_\phi$: (1) future state tokens $\mathbf{z}_{\text{future}}$, and (2) value estimates $v_t$. Future state tokens are projected via a lightweight MLP to align their dimensionality with the policy's visual encoder outputs. Value estimates are transformed into action advantages using $n$-step temporal difference estimation:

\begin{equation}
    A(\mathbf{s}_t, a_t) = \sum_{k=0}^{n-1} \gamma^k r_{t+k} + \gamma^n v_{t+n} - v_t,
\end{equation}

\noindent where $v_t$ and $v_{t+n}$ denote value predictions for states $\mathbf{s}_t$ and $\mathbf{s}_{t+n}$ respectively, and $\gamma$ is the discount factor. To simplify conditioning while preserving preference structure, advantages are discretized into a binary indicator $I = \mathds{1}\big(A(\mathbf{s}_t, a_t) > \epsilon\big)$ with threshold $\epsilon$. The policy is then trained to generate actions conditioned on the tuple $(I, \mathbf{z})$ by minimizing the supervised fine-tuning objective defined in Eq.~\ref{eq:sft}.

To prevent over-reliance on synthetic world model signals and ensure flexible deployment, we adopt two complementary strategies during training. First, the world model performs only a single denoising step during inference to minimize computational overhead. Second, we implement stochastic attention masking that randomly suppresses world model tokens with probability $p=0.2$ during training. This forces the policy to maintain robust performance even when world model inputs are partially or fully unavailable, enabling an efficient inference mode that bypasses world model conditioning.

\textbf{Stage 3: Human-in-the-Loop Rollout Data Collection.} In the third phase, we deploy the policy to collect trajectories through human-in-the-loop rollouts. The resulting dataset comprises a hybrid mixture of autonomous executions and expert interventions. Autonomous rollouts exhibit a significantly reduced action distribution gap compared to conventional teleoperation, since the policy generates actions in its native distribution rather than mimicking human demonstrations, thereby providing more effective supervision signals for VLA learning. However, autonomous execution inevitably encounters failure modes requiring human correction. To mitigate temporal discontinuities introduced by manual interventions, we developed a human-in-the-loop rollout data collection software that automatically detects and removes transitional artifacts at intervention boundaries. This smoothing mechanism ensures temporal coherence across the entire trajectory, producing a clean, continuous dataset that facilitates stable policy updates in the subsequent training stage while preserving the pedagogical value of expert corrections.

\textbf{Stage 4: Continual Training with Rollout Data.} In this stage, we fine-tune the policy using the curated HILR dataset to master complex long-horizon behaviors emerging from the diverse mixture of autonomous executions and expert corrections. Crucially, to prevent advantages collapse toward zero ($A(\mathbf{s}_t, a_t) \approx 0$), the world model $\mathcal{W}_\phi$ is jointly trained with HILR dataset and base data.
For the policy training, consistent with Stage 2, we maintain stochastic attention masking with masking probability $p=0.2$ applied to both the advantage indicator $I$ and future latent tokens $\mathbf{z}_{\text{future}}$. This regularization serves dual purposes: (1) it prevents policy over-reliance on world model signals by forcing robustness to missing conditioning inputs, and (2) it ensures architectural and training consistency between pretraining and fine-tuning phases, avoiding distributional shift at inference time.
The rollout–annotation–training cycle operates iteratively, establishing a self-improving closed loop: as the policy improves, its autonomous rollouts cover increasingly complex and successful behaviors, which in turn generate higher-quality training data for subsequent iterations.

\textbf{Inference.} During deployment, we enforce an optimistic control strategy by fixing the advantage indicator to $I=1$. 
Regarding the latent condition $\mathbf{z}$, the architectural decoupling facilitated by stochastic masking enables two flexible execution modes: (1) an efficient mode, where the world model is bypassed to maximize inference frequency. In this setting, the attention mask is configured to render the future latent tokens invisible to the policy, compelling it to act based solely on current observations; (2) and a standard mode, where the world model actively generates $\mathbf{z}$ to provide dense look-ahead guidance. In this setting, the attention mask permits full visibility of the predicted future states, allowing the policy to leverage the prospective context for complex, long-horizon planning.

\section{Experiment}

In this section, we first evaluate the performance of our foundation model, \textit{GigaBrain-0.5}. On internal robotic evaluation, our model demonstrates robust capability in executing long-horizon, complex procedures such as box packing and coffee preparation. On the public benchmark RoboChallenge~\citep{robochallenge}, our foundation model also achieves superior performance compared to $\pi_{0.5}$~\citep{pi05}. 
Next, we compare our world model-based reinforcement learning approach, \textit{RAMP}, against established RL baselines including AWR~\citep{awr} and RECAP~\cite{pi06}. Experimental results confirm that \textit{RAMP} exhibits significantly higher sample efficiency and stronger multi-task generalization capability.
Finally, we conduct ablation studies to analyze the contribution of the value prediction module within our world model, quantitatively validating its importance for policy learning and task success.

\subsection{Foundation Model Performance}

\noindent
\label{sec:data}
\textbf{Pre-training Details.}
\textit{GigaBrain-0.5} is pre-trained on a diverse dataset exceeding 10,000 hours, comprising over 6,000 hours of world model-generated data and approximately 4,000 hours of real-robot collected data. The detailed data composition is illustrated in Fig.~\ref{fig:pretraindata}. We train \textit{GigaBrain-0.5} using our training framework GigaTrain\footnote{\url{https://github.com/open-gigaai/giga-train}} with a batch size of 3,072 for 100,000 optimization steps. To reduce per-GPU memory consumption, we employ Fully Sharded Data Parallel (FSDP) v2, applying sharding selectively to all \texttt{SiglipEncoderLayer} modules and only the first 16 layers of \texttt{Gemma2DecoderLayerWithExpert}.

\textbf{Post-training Details.}
To evaluate the performance of \textit{GigaBrain-0.5} on physical robots, we collect task-specific demonstration data on the target robot platform and perform post-training to adapt the model to each task. We conduct comprehensive evaluations on eight internally designed tasks and additionally post-trained the model on 30 tasks from the public benchmark RoboChallenge. Details of the RoboChallenge tasks and evaluation protocol are described in~\citep{robochallenge}. Our eight internal evaluation tasks include \texttt{Juice Preparation}, \texttt{Box Moving}, \texttt{Table Bussing}, \texttt{Paper Towel Preparation}, \texttt{Laundry Folding}, \texttt{Laundry Collection}, \texttt{Box Packing}, and \texttt{Espresso Preparation}. Demonstration videos for these tasks are showcased on our project page. For each task, we performed post-training with a batch size of 256 for 20,000 optimization steps.

\begin{figure}[t]
\centering
\captionsetup{type=figure, justification=justified, singlelinecheck=false}
\includegraphics[width=1\linewidth]{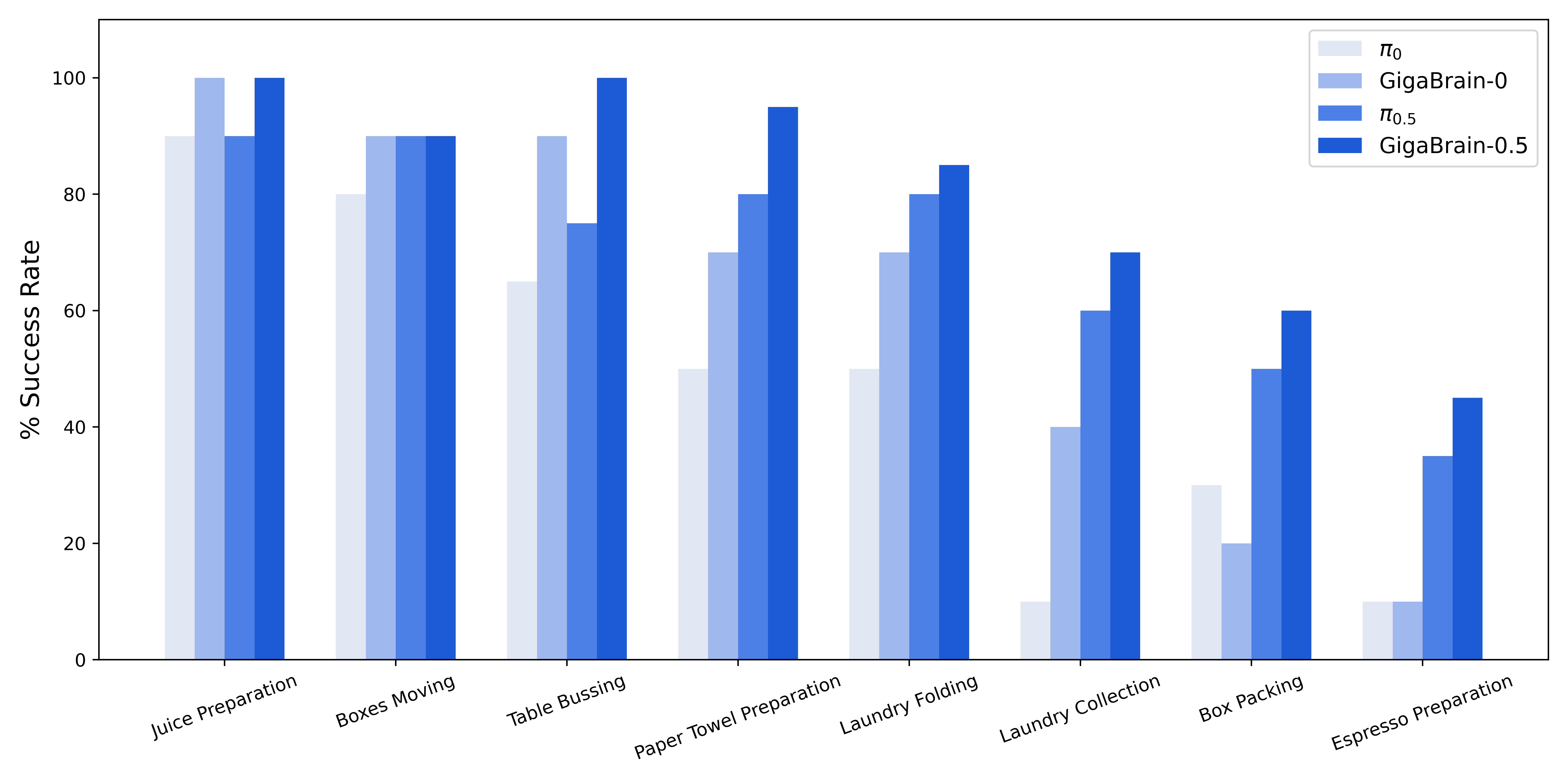}
\caption{Performance of \textit{GigaBrain-0.5} on internal evaluation.}
\label{fig:gigabrain05_exp}
\end{figure}

\begin{figure}[!t]
\centering
\captionsetup{type=figure, justification=justified, singlelinecheck=false}
\includegraphics[width=\linewidth]{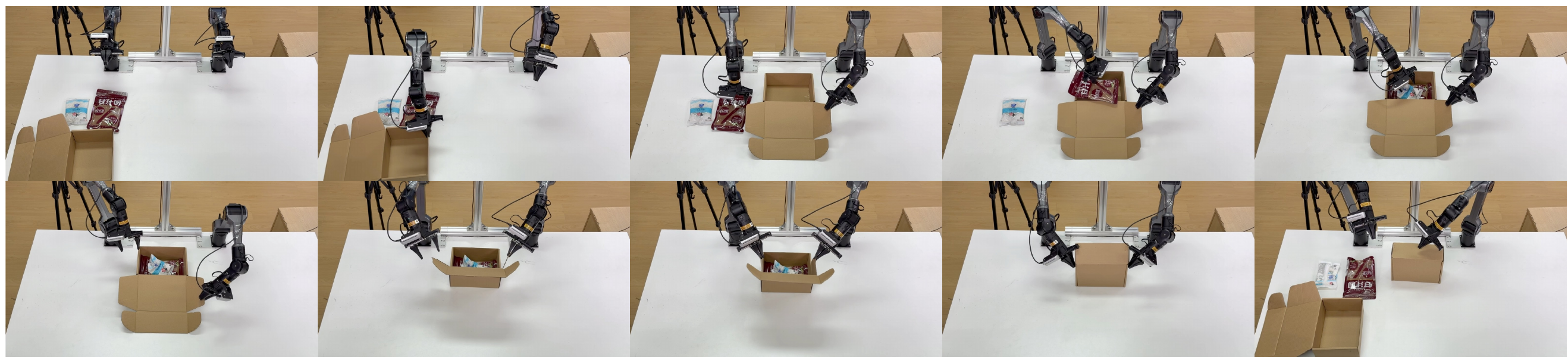}
\caption{Deployment of \textit{GigaBrain-0.5} on PiPER arms for real-world \texttt{Box Packing}.}
\label{fig:demos_box_packing}
\end{figure}

\begin{figure}[!t]
\centering
\captionsetup{type=figure, justification=justified, singlelinecheck=false}
\includegraphics[width=\linewidth]{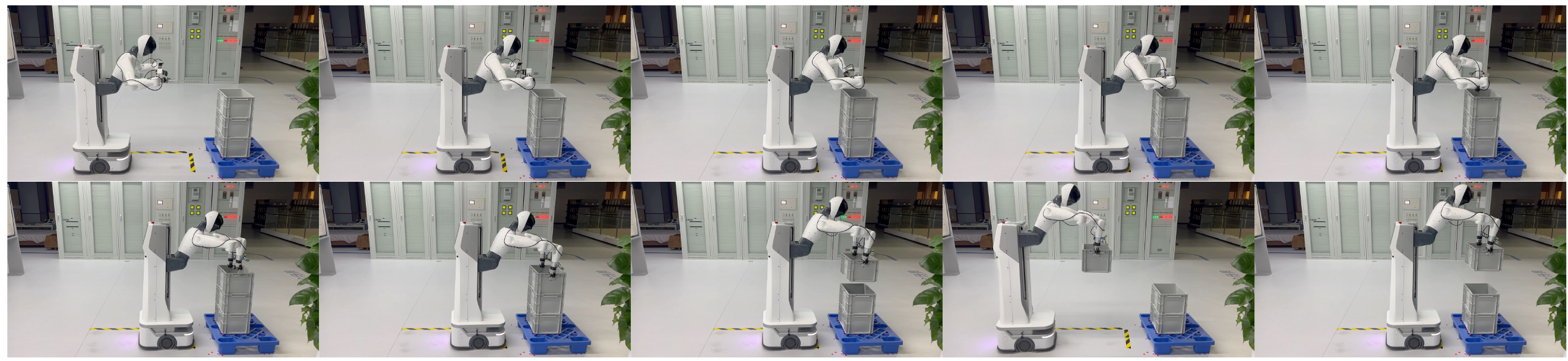}
\caption{Deployment of \textit{GigaBrain-0.5} on the G1 humanoid robot for real-world \texttt{Box Moving}.}
\label{fig:demos_boxes_moving}
\end{figure}

\begin{figure}[!t]
\centering
\captionsetup{type=figure, justification=justified, singlelinecheck=false}
\includegraphics[width=\linewidth]{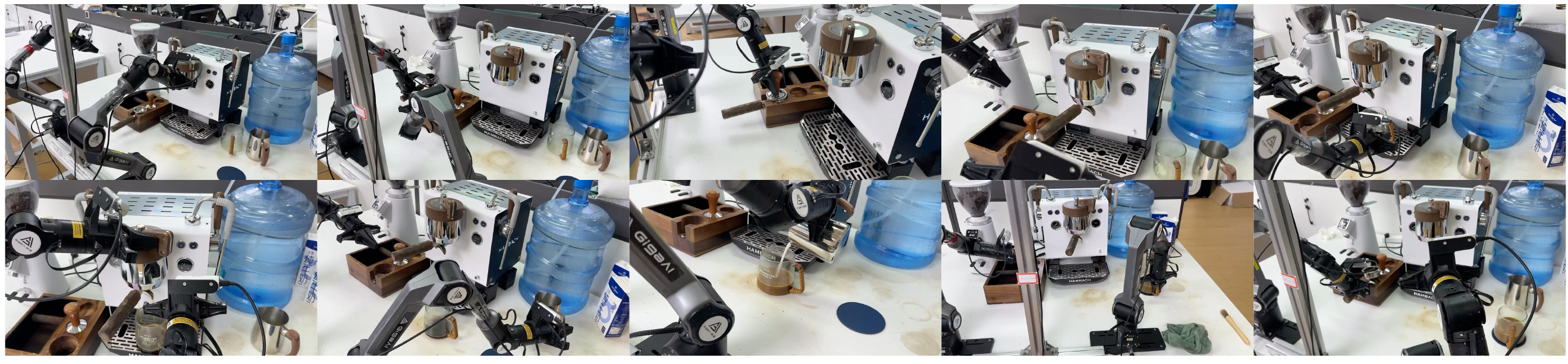}
\caption{Deployment of \textit{GigaBrain-0.5} on PiPER arms for real-world \texttt{Espresso Preparation}.}
\label{fig:demos_espresso_preparation}
\end{figure}

\begin{figure}[!t]
\centering
\captionsetup{type=figure, justification=justified, singlelinecheck=false}
\includegraphics[width=\linewidth]{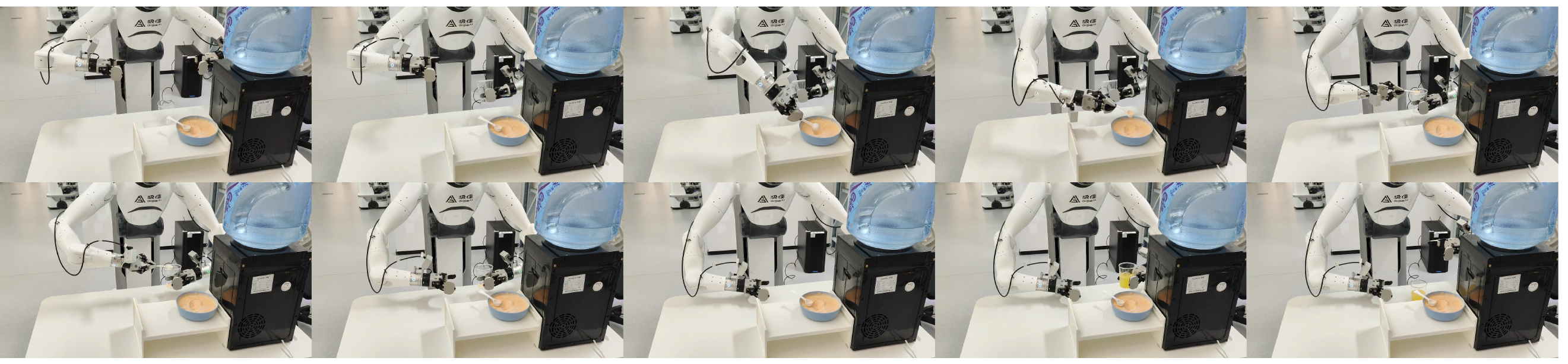}
\caption{Deployment of \textit{GigaBrain-0.5} on the G1 humanoid robot for real-world \texttt{Juice Preparation}.}
\label{fig:demos_juice_preparation}
\end{figure}

\begin{figure}[!t]
\centering
\captionsetup{type=figure, justification=justified, singlelinecheck=false}
\includegraphics[width=\linewidth]{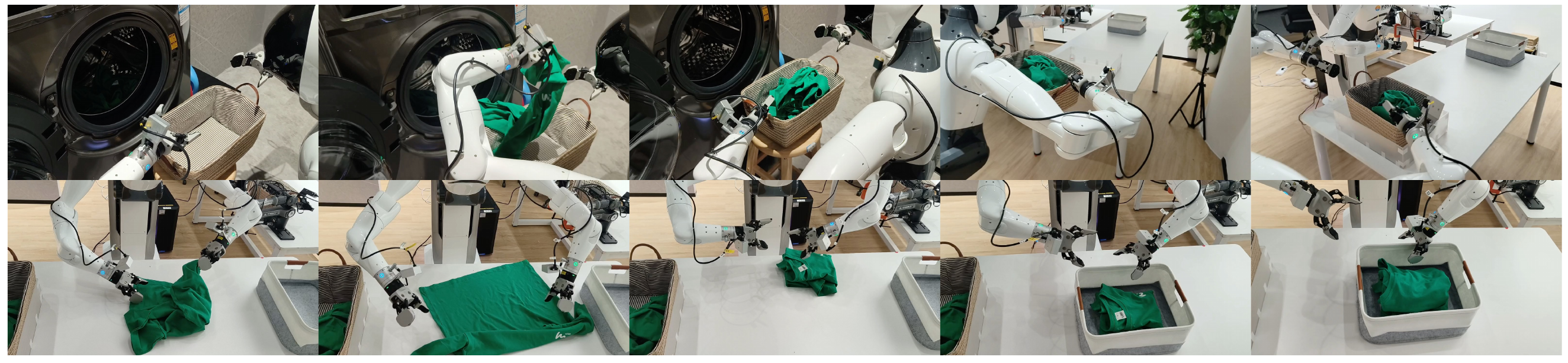}
\caption{Deployment of \textit{GigaBrain-0.5} on the G1 humanoid robot for real-world \texttt{Laundry Collection}.}
\label{fig:demos_laundry_collection}
\end{figure}

\begin{figure}[!t]
\centering
\captionsetup{type=figure, justification=justified, singlelinecheck=false}
\includegraphics[width=\linewidth]{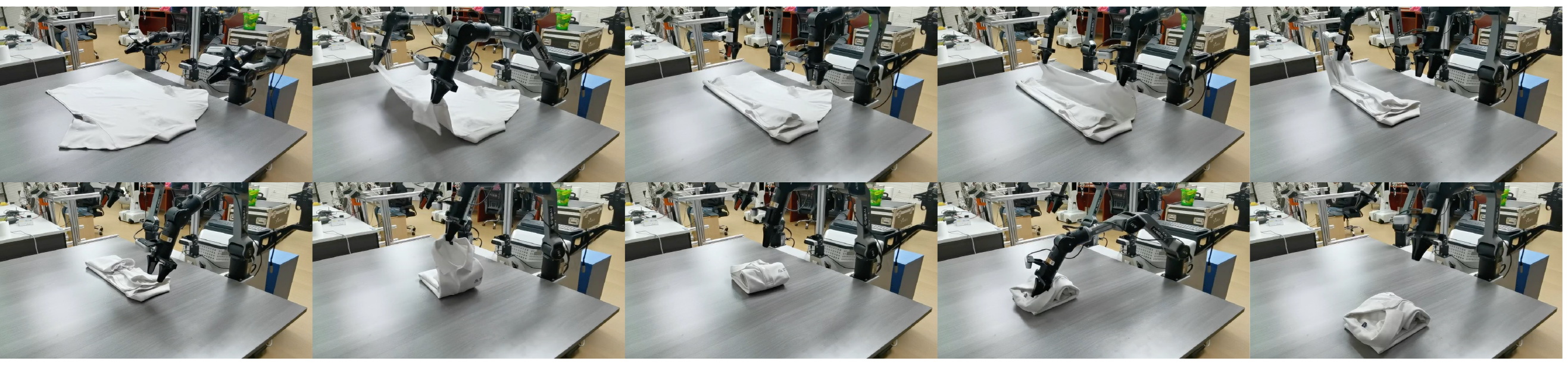}
\caption{Deployment of \textit{GigaBrain-0.5} on PiPER arms for real-world \texttt{Laundry Folding}.}
\label{fig:demos_laundry_folding}
\end{figure}

\begin{figure}[!t]
\centering
\captionsetup{type=figure, justification=justified, singlelinecheck=false}
\includegraphics[width=\linewidth]{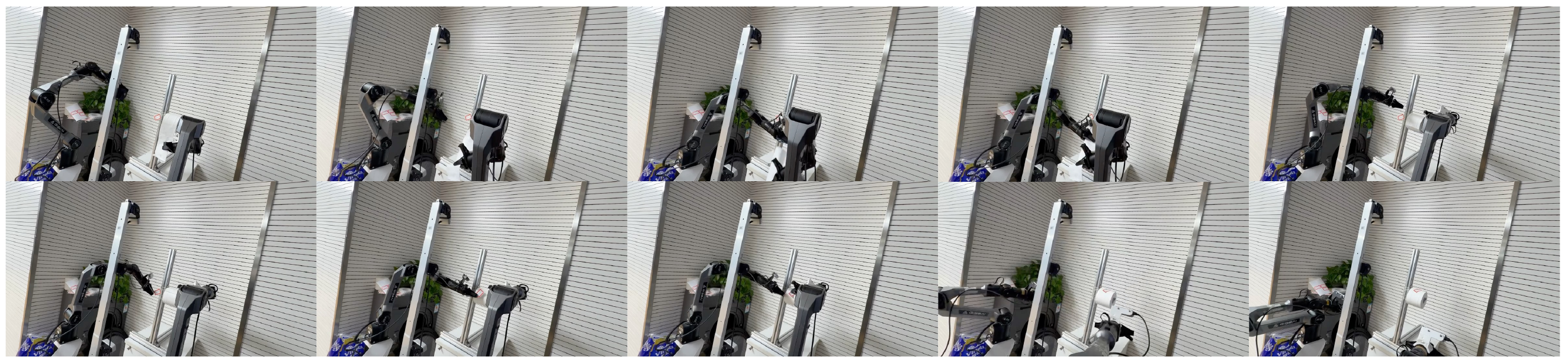}
\caption{Deployment of \textit{GigaBrain-0.5} on PiPER arms for real-world \texttt{Paper Towel Preparation}.}
\label{fig:demos_paper_towel_preparation}
\end{figure}

\begin{figure}[!t]
\centering
\captionsetup{type=figure, justification=justified, singlelinecheck=false}
\includegraphics[width=\linewidth]{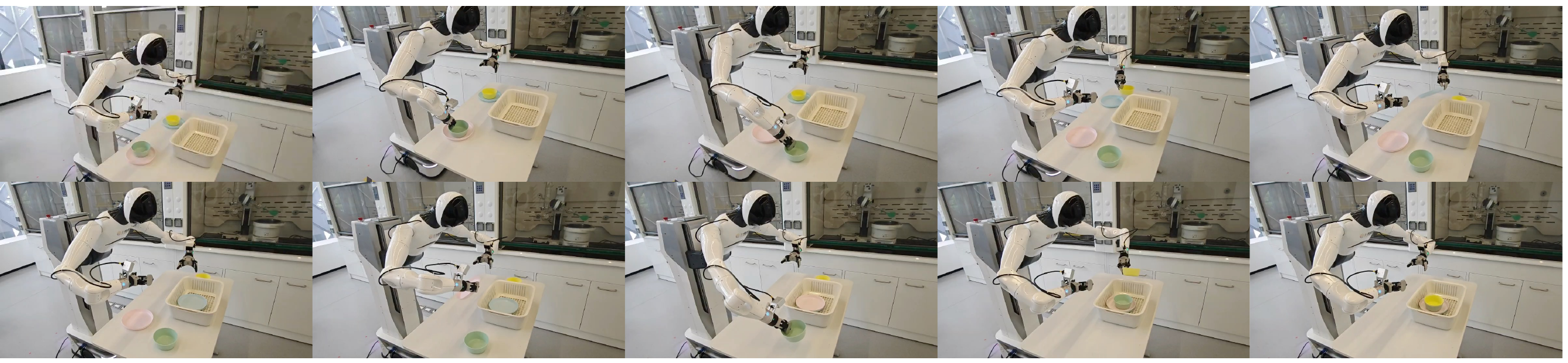}
\caption{Deployment of \textit{GigaBrain-0.5} on the G1 humanoid robot for real-world \texttt{Table Bussing}.}
\label{fig:demos_table_bussing}
\end{figure}

\noindent
\textbf{Internal Evaluation.} In our experiments, we benchmark \textit{GigaBrain-0.5} against several strong baselines, including $\pi_0$~\citep{pi0}, $\pi_{0.5}$, and GigaBrain-0~\citep{gigabrain0}. The results are summarized in Fig.~\ref{fig:gigabrain05_exp}. \textit{GigaBrain-0.5} achieves consistent and substantial improvements over its predecessor GigaBrain-0~\citep{gigabrain0} across all evaluated tasks, attaining the highest success rate in every case, with particularly notable gains in complex multi-step procedures. For instance, in \texttt{Juice Preparation}, a task requiring sequential ingredient handling and mixing, \textit{GigaBrain-0.5} achieves a 100\% success rate, surpassing GigaBrain-0's 90\%. For challenging tasks such as \texttt{Box Packing} and \texttt{Espresso Preparation}, \textit{GigaBrain-0.5} improves success rates by 10\% and 20\% respectively over $\pi_{0.5}$. Similarly, on highly dexterous manipulation tasks (\texttt{Paper Towel Preparation}, \texttt{Laundry Folding}, and \texttt{Laundry Collection}), \textit{GigaBrain-0.5} achieves success rates exceeding 80\%, outperforming $\pi_{0.5}$ by 15\%, 5\%, and 10\% respectively. Additionally, we visualize eight tasks in Fig.~\ref{fig:demos_box_packing}-Fig.~\ref{fig:demos_table_bussing}.

\noindent
\textbf{RoboChallenge Evaluation.} Beyond our internal task evaluations, we also conduct comprehensive assessments on the RoboChallenge~\citep{robochallenge} benchmark. RoboChallenge represents the world's first large-scale embodied AI evaluation platform featuring real-robot testing. The platform has established standardized remote evaluation protocols across a cluster of 20 physical robots spanning four major platforms (UR5, Franka, ARX5, and ALOHA). It further provides the open-sourced dataset (736\,GB) encompassing 30 standardized manipulation tasks. Detailed task specifications and evaluation methodologies are provided in~\citep{robochallenge}. An intermediate iteration model (GigaBrain-0.1) currently ranks first on the leaderboard as of February 9, 2026, achieving an average success rate of 51.67\%, an improvement of 9\% over $\pi_{0.5}$ (42.67\%).

\subsection{RAMP Performance}
\label{sec:exp_RAMP}

In this section, we address three core questions through systematic empirical evaluation: (1) Does world model-based value prediction offer superior accuracy and efficiency compared to the VLM-based approach employed in $\pi^*_{0.6}$? (2) Does world model conditioning enhance cross-task generalization capabilities of vision-language-action policies? (3) How does our proposed \textit{RAMP} algorithm compare against alternative reinforcement learning methods in real-robot settings?

\noindent \textbf{Value Prediction Performance.} 
To evaluate the efficacy of world model-based value prediction, we conduct comparisons against a VLM-based baseline~\citep{pi06} and our world model-based method. For value prediction in the VLM, we insert a learnable \texttt{[CLS]} token at the end of the visual token sequence to aggregate global scene representations, the hidden state of this token is then projected through a regression head to produce a scalar value prediction in $[0, 1]$. The model is optimized with mean squared error.
Both VLM-based and world model-based value predictors are trained on identical pretraining data and evaluated on a validation set comprising approximately 1 million frames across the eight manipulation tasks illustrated in Fig.~\ref{fig:gigabrain05_exp}. We assess prediction quality using four complementary metrics: MAE, MSE, RMSE (all lower is better), and Kendall's tau rank correlation coefficient (higher is better, with 1 indicating perfect rank preservation). Results averaged across all tasks are summarized in Tab.~\ref{tab:reward_comparison}.

\begin{figure}[!t]
\centering
\captionsetup{type=figure, justification=justified, singlelinecheck=false}
\includegraphics[width=1\linewidth]{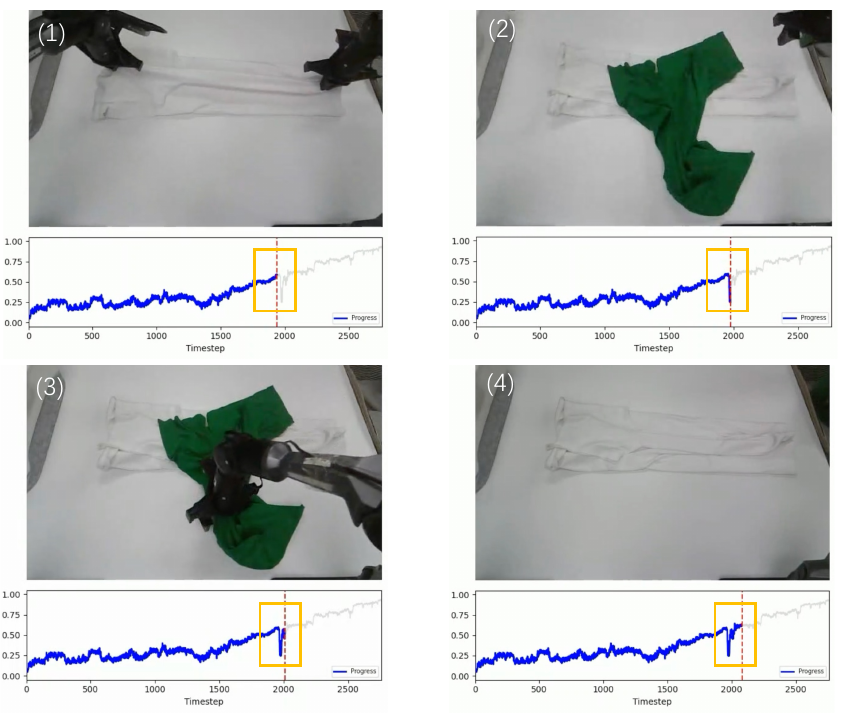}
\caption{Value prediction visualization from the world model. The \textcolor{orange}{orange} bounding box highlights a value drop during a \texttt{Laundry Folding} task when a green garment interferes with the folding process, the predicted value recovers after the manipulator successfully removes the obstruction.}
\label{fig:reward}
\end{figure}

Our analysis reveals three key findings. First, despite employing a lightweight VLM~\citep{pi06}, the VLM-based approach incurs the highest per-frame inference latency (0.32\,s, on A800 GPU) due to the computational overhead of the SigLIP~\citep{siglip} visual encoder. Second, the world model variant predicting value alone achieves the fastest inference (0.11\,s) but suffers from degraded prediction accuracy (MAE=0.0838, Kendall=0.7288), suggesting that value-only modeling fails to fully exploit the future prediction capabilities inherent in world models. Third, our proposed joint prediction scheme, simultaneously forecasting value and future states, strikes an optimal balance. It achieves the highest Kendall's tau (0.8018) and lowest MAE (0.0621) while maintaining competitive inference speed (0.25\,s). This demonstrates that leveraging future state prediction provides crucial contextual grounding for accurate value estimation. Qualitative value prediction visualizations for representative tasks are provided in Fig.~\ref{fig:reward}.

\begin{table}[h]
\centering
\vspace{-0.5em}
\caption{Performance comparison of different value prediction methods.}
\vspace{-0.5em}
\label{tab:reward_comparison}
\begin{tabular}{l c c c c c}
\toprule
\textbf{Model} &  \textbf{Inference Time (s)} & \textbf{MAE} $\downarrow$& \textbf{MSE}$\downarrow$ & \textbf{RMSE} $\downarrow$ & \textbf{Kendall}$\uparrow$  \\
\midrule
VLM-based & 0.32 & 0.0683 & 0.0106 & 0.1029 & 0.7972 \\
\midrule
WM-based (value only) & 0.11 & 0.0838 & 0.0236 & 0.1433 & 0.7288 \\
\midrule
WM-based (state+value) & 0.25 & 0.0621 & 0.0099 & 0.0989 & 0.8018 \\
\bottomrule
\end{tabular}
\end{table}

\begin{figure}[htbp]
\centering
\captionsetup{type=figure, justification=justified, singlelinecheck=false}
\includegraphics[width=1\linewidth]{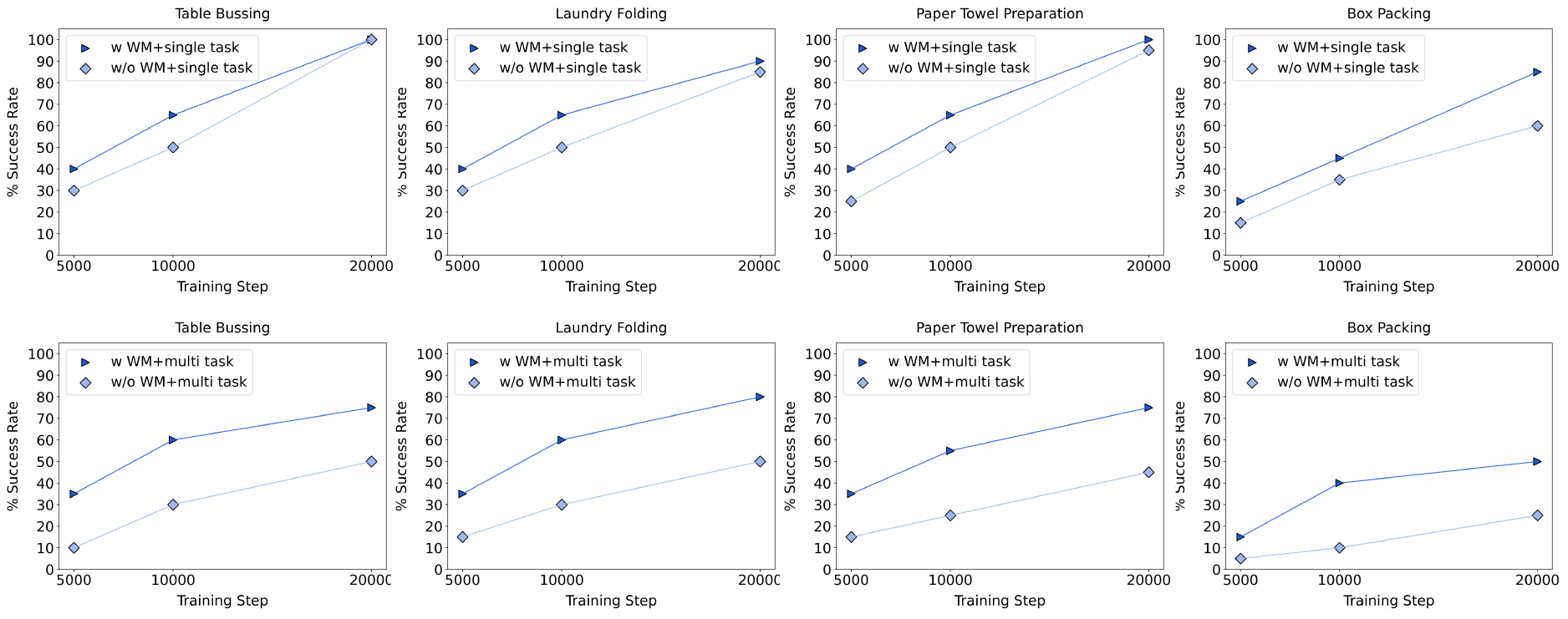}
\caption{Comparison of single-task and multi-task performance with and without world model conditions.}
\vspace{-0.5em}
\label{fig:multi_task_exp}
\end{figure}

\noindent \textbf{World Model Conditioning for Policy Learning.}
To evaluate whether world model conditioning enhances multi-task generalization, we conduct a controlled comparison between single-task and multi-task training regimes. We select four representative manipulation tasks for evaluation: \texttt{Table Bussing}, \texttt{Laundry Folding}, \texttt{Paper Towel Preparation}, and \texttt{Box Packing}. To ensure a fair comparison and isolate the effect of world model conditioning, we train all policies exclusively on the Stage-2 dataset from \textit{RAMP} without incorporating any generated rollout data.
For single-task training, each policy is trained independently for 20000 steps with a batch size of 256. For multi-task training, we uniformly mix data from all four tasks and train a single policy for 60000 steps using the same batch size. 
Our experimental results, as illustrated in Figure~\ref{fig:multi_task_exp}, demonstrate that the world model condition approach consistently outperforms the baseline across both single-task and multi-task training scenarios. Specifically, incorporating the world model yields substantial performance gains in all evaluated tasks, with significant improvements consistently observed throughout the entire training trajectory from 5,000 to 20,000 steps. Notably, the performance enhancement is particularly pronounced in the multi-task setting, where the success rate gap between the world model approach and the baseline widens progressively during training, achieving up to $\sim$30\% higher success rates in tasks like \texttt{Box Packing} at step 20000. This indicates that the world model condition effectively facilitates knowledge transfer across multiple tasks while maintaining robust performance gains in single-task scenarios.

\noindent \textbf{Comparison with RL Baselines.}
We benchmark \textit{RAMP} against state-of-the-art RL baselines.
\begin{itemize}
    \item \textit{GigaBrain-0.5} + \textbf{AWR}~\citep{awr}: An offline RL baseline that fine-tunes the \textit{GigaBrain-0.5} policy using weighted imitation learning, utilizing rollouts generated by the current policy.
    \item \textit{GigaBrain-0.5} + \textbf{RECAP}~\citep{pi06}: An advantage-conditioned offline RL approach that extends the \textit{GigaBrain-0.5} backbone with an advantage input, serving as an ablated variant of our method without state prediction.
    \item \textit{GigaBrain-0.5} + \textbf{RAMP} (\textit{GigaBrain-0.5M*}): The proposed Reinforcement learning via world Model-conditioned Policy framework, which conditions the \textit{GigaBrain-0.5} policy on both the predicted value and future state latents to optimize long-horizon task performance.
\end{itemize}

\begin{figure}[htbp]
\centering
\captionsetup{type=figure, justification=justified, singlelinecheck=false}
\includegraphics[width=1\linewidth]{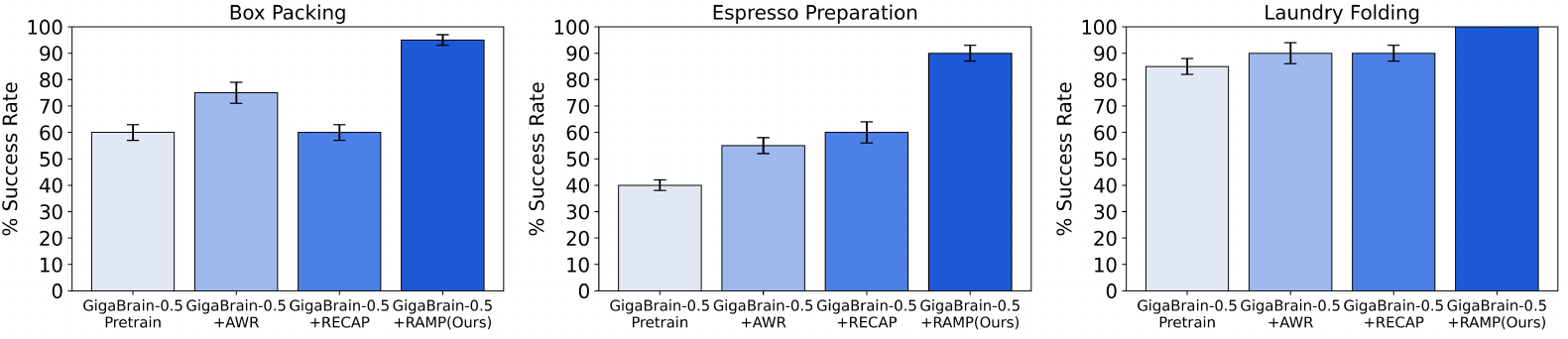}
\caption{Comparison of different RL methods.}
\label{fig:comp_rl_exp}
\end{figure}

Our \textit{RAMP} framework establishes top performance across all three highly challenging manipulation tasks: \texttt{Box Packing}, \texttt{Espresso Preparation}, and \texttt{Laundry Folding}. As quantified in Fig.~\ref{fig:comp_rl_exp}, \textit{RAMP} achieves near-perfect success rates on all evaluated tasks, significantly outperforming all baseline methods (\textit{GigaBrain-0.5}, \textit{GigaBrain-0.5}+AWR, and \textit{GigaBrain-0.5}+RECAP). Notably, \textit{RAMP} demonstrates particularly substantial gains on \texttt{Box Packing} and \texttt{Espresso Preparation}, where it surpasses the RECAP baseline by approximately 30\% points. 
Critically, the \textit{GigaBrain-0.5M$^*$} model (i.e., \textit{GigaBrain-0.5} integrated with our \textit{RAMP} framework) exhibits robust and consistent task execution capabilities, achieving reliable success in real-world deployment as empirically validated by the supplementary execution videos on our project page. This unprecedented performance across multiple complex manipulation tasks underscores the effectiveness of \textit{RAMP} in solving challenging real-world robotics problems.

\section{Conclusion and Future Work}
\label{sec:conclusion}

In this work, we present \textit{GigaBrain-0.5} and its world model-enhanced successor \textit{GigaBrain-0.5M*}, advancing the frontier of VLA learning through large-scale pretraining and model-based RL. \textit{GigaBrain-0.5}, pretrained on over 10,000 hours of diverse robotic data, demonstrates state-of-the-art performance across eight internal manipulation tasks and 30 standardized tasks on the RoboChallenge benchmark, achieving a 51.67\% average success rate and securing the top position on the public leaderboard. Building upon this strong foundation, \textit{GigaBrain-0.5M*} introduces a novel world model-conditioned architecture that leverages future state prediction to overcome the limited anticipation capabilities inherent in conventional VLA models. By integrating model-based reinforcement learning through \textit{RAMP}, our approach achieves robust cross-task generalization and reliably executes complex long-horizon tasks such as sequential box packing and espresso preparation. Looking ahead, the \textit{GigaBrain} series will investigate more efficient utilization of model rollout data to maximize the informational value of synthetic trajectories while minimizing computational overhead. Furthermore, we aim to explore more scalable self-evolution paradigms that enable autonomous data curation, policy refinement, and world model updating through closed-loop interaction.

\clearpage
\setcitestyle{numbers}
\bibliographystyle{plainnat}
\bibliography{main}

\end{document}